\documentclass[conference]{IEEEtran}
\IEEEoverridecommandlockouts
\usepackage{cite}
\usepackage{amsmath,amssymb,amsfonts}
\usepackage{algorithmic}
\usepackage{graphicx}
\usepackage{textcomp}
\usepackage{xcolor}
\usepackage{tabularx}
\usepackage{multirow}
\usepackage{subfig}
\usepackage{xcolor}
\usepackage{soul}
\usepackage{caption}
\DeclareCaptionFont{eightpt}{\fontsize{8pt}{11pt}\selectfont #1}
\begin{document}

\title{The Effectiveness of Temporal Dependency in Deepfake Video Detection\\

}

\author{\IEEEauthorblockN{Will Rowan and Nick Pears}
\IEEEauthorblockA{Department of Computer Science \\
University of York\\
York, United Kingdom \\
{wjr508@york.ac.uk, nick.pears@york.ac.uk}
}}

\maketitle

\begin{abstract}

Deepfakes are a form of synthetic image generation used to generate fake videos of individuals for malicious purposes. The resulting videos may be used to spread misinformation, reduce trust in media, or as a form of blackmail. These threats necessitate automated methods of deepfake video detection. This paper investigates whether temporal information can improve the deepfake detection performance of deep learning models.

To investigate this, we propose a framework that classifies new and existing approaches by their defining characteristics. These are the types of feature extraction: automatic or manual, and the temporal relationship between frames: dependent or independent. We apply this framework to investigate the effect of temporal dependency on a model's deepfake detection performance.

We find that temporal dependency produces a statistically significant (p $<$ 0.05) increase in performance in classifying real images for the model using automatic feature selection, demonstrating that spatio-temporal information can increase the performance of deepfake video detection models.

\end{abstract}

\section{Introduction}

Recently, generative machine learning techniques have enabled the creation of synthetic, photo-realistic videos that appear to show real people saying and doing things they have never said or done. The resulting videos have become known as \emph{deepfakes} — recognising the deep learning techniques used in their creation and their fake nature. We define deepfakes as a form of video manipulation that involves swapping the faces of individuals using deep learning techniques. This is distinguished from traditional methods relying on conventional video editing techniques that are known as \emph{cheapfakes} \cite{paris2019deepfakescheapfakes}. We further distinguish deepfakes from other forms of synthetic image generation by their content and purpose: deepfakes are synthetic images of the human face designed for malicious purposes.

Image editing techniques have been available to the general public for many years. However, editing videos using traditional means, such as post-processing ,is time consuming and costly — requiring expertise or significant resources to produce convincing fake videos. The prevalence and dangers of deepfakes are due the ease with which they can now be generated using commodity GPUs. Taking advantage of recent advances in deep learning, deepfakes have become increasingly realistic and easy to produce. GUI solutions such as Doublicat \cite{doublicat} allow non-technical users to experiment with similar techniques on mobile devices. Social media platforms allow for such videos to reach a large audience.

These advances in synthetic image generation raise numerous legal, social, and ethical problems. State-of-the-art facial recognition systems such as FaceNet \cite{schroff2015facenet} have been shown to be highly susceptible to deepfakes \cite{korshunov2018deepfakesdftimit}. Political deepfakes can be used to spread misinformation, reducing trust in reputable media \cite{vaccari2020deepfakes}. Early deepfake solutions were exploited to create non-consensual deepfake pornography \cite{chesneyposneg2019deep}. Clearly, there is a need for accurate methods of deepfake video detection.

State of the art deepfake detection approaches utilise deep learning to predict the probability of an image or video being a deepfake. They have shown to be accurate on earlier, less diverse datasets but struggle to generalise in more challenging conditions — where there are a large number of individuals exhibiting various poses in an unconstrained setting. The winning model in the Deepfake Detection Challenge (DFDC) achieved an average precision of 65.18\% \cite{dolhansky2020deepfakedfdc}. This demonstrates that small datasets do not capture enough diversity to train models that generalise well \emph{in-the-wild}.

Approaches to deepfake detection vary widely in their implementation: the input they accept and the processes applied to this input. Whereas deepfake datasets have been classified by their characteristics \cite{li2019celeb}, deepfake detection methods have not. To help quantify the effect of temporal dependency on performance, we present a novel framework which isolates temporal dependency as a characteristic. The proposed framework classifies deepfake detection techniques using two criteria: the method of feature extraction and the temporal relationship between frames. We adapt an existing deepfake detection model for each of the identified types of system; comparing their results on a benchmark dataset. Their precision on several deepfake generation methods is evaluated and tested for statistical significance. We show clear trade-offs between these types of detection techniques and make suggestions on where further research in the area should focus.

\section{Related Work}

\subsection{Deepfake Generation Methods}

Facial manipulation methods vary by purpose and implementation but all result in the manipulation of an individual in a target video. Techniques include face replacement \cite{nirkin2019fsgan}, visual dubbing where an audio track is used to produce lip-synced video \cite{suwajanakorn2017synthesizingobama}, and facial re-enactment \cite{teadies2016face2face}. Both facial re-enactment and face replacement take features from a source video and apply them to a target video. This results in a manipulated target video. Deepfakes are the result of deep learning being applied to the task of facial replacement \cite{rossler2019faceforensics++xception}. 

Facial replacement can be performed using a variety of techniques, including 3D morphable models, generative adversarial networks, and autoencoders. However, the majority of deepfakes are created using open-source implementation such as DeepFaceLab \cite{petrov2020deepfacelab} which use an encoder-decoder architecture. The deepfake algorithm seeks to transform instances of one individual in a video into another individual using their respective facial features, including their expression. To do so, the algorithm requires a large number of training examples of each individual. The process can be broken down into three sequential steps: face extraction and alignment, training the auto-encoder, and face warping. This is repeated for every frame in the target video.

\subsection{Deepfake Datasets}

Deepfake detection datasets allow for the training and evaluation of a deepfake detection system. Training examples should be in sufficient number and variety to allow a model to generalise well to unseen examples. For evaluation, the dataset should include examples of deepfakes and real videos as would be found \emph{in-the-wild} — being equivalent to the setting in which such a deepfake detection model would be deployed. 

Li et al. \cite{li2019celeb} categorise existing datasets into two distinct generations based on the time of publication and the synthesis algorithms used in their deepfake generation processes. Dolhansky et al. \cite{dolhansky2020deepfakedfdc} go further in proposing a third generation of techniques;  these have consent from their participants in addition to containing an order of magnitude larger corpus of higher quality frames and videos compared to the second generation. These third generation datasets include a much larger number of individuals — reducing the likelihood of models overfitting on individuals present within the dataset.

\subsection{Deepfake Detection Methods}

Four types of deepfake detection methods will now be considered in turn — these represent the state of the art in deepfake video detection.

\subsubsection{Generation Artefact Approaches}

These detection methods focus on detecting artefacts from all three stages of the deepfake generation process. Yang et al. \cite{yang2019exposingheadpose} use inconsistencies in the alignment of the face interior, the part synthesised in the deepfake algorithm, and outer facial landmarks of an individual as a feature to detect deepfakes. Yang et al. \cite{yang2019exposingheadpose} also use the difference in the estimated head pose of the face when using landmarks from the central face region and the whole face. Alignment errors in the splicing process result in this being a distinguishing feature. 

Li et al. \cite{li2018exposingfwa} exploit the face warping stage of the deepfake generation pipeline, using the ResNet50 \cite{he2016deepresnet} architecture to detect deepfakes. The artefacts produced by the deepfake pipeline can be simulated on standard facial datasets, meaning examples of deepfakes are not required for training. Each of these outlined approaches uses a manually selected feature designed to exploit a weakness in the deepfake generation process. They all make predictions on a single frame, not taking advantage of temporal information across a video.

\subsubsection{Physiological Signals}

Using techniques from the closely-related field of media forensics, physiological signals of human behaviour can be used to detected deepfakes. Li et al. \cite{li2018ictueyes} use a CNN and LSTM in combination to predict when an individual blinks in a given video. It is suggested that this manually selected feature can use this temporal information to discriminate between deepfakes and real videos. However, results using the technique have yet to be presented.

\subsubsection{Data-Driven Approaches}

Data-driven approaches use deep neural networks to detect deepfake videos. Rossler et al. \cite{rossler2019faceforensics++xception} use transfer learning to advance the state of the art in data-driven deepfake detection.  They repurpose weights learnt from the ImageNet \cite{deng2009imagenet} dataset as a starting point for XceptionNet \cite{chollet2017xception}, training it on the FaceForensics++ dataset to detect deepfakes. The resulting approach is the best performer on the state of the art FaceForensics++ dataset \cite{rossler2019faceforensics++xception} as is a variant of the same network on the Celeb-DF dataset \cite{li2019celeb}. Both implementations use automated feature selection without temporal information to detect deepfakes.

\subsubsection{Other Methods}

Social video verification verifies the truth of an event from multiple videos at different viewpoints \cite{tursman2020towardssocialvideoveri}. Tursman et al. \cite{tursman2020towardssocialvideoveri} propose a technique which compares the facial geometry of the subject across the videos — determining for each video whether it is a deepfake or not.

\subsection{Biases within Deepfake Detection}

Due to the large number of images required to train a deepfake model, deepfake datasets typically consist of individuals within the public eye. For example, Celeb-DF \cite{li2019celeb} comprises of images of celebrities. This tendency leaves deepfake detection vulnerable to replicating existing systemic biases in the fields they draw data from. Unless corrected for, unequal representation in those who are celebrities will be replicated throughout the deepfake detection pipeline. 

 Deepfake detection models should be evaluated using a dataset balanced by gender and race. The results should be published for transparency, allowing for public trust in the system. Commercial gender classification systems have been shown to disproportionately misclassify subgroups of individuals \cite{buolamwini2018gender}. It is recognised that this bias stemmed from datasets which are overwhelmingly white and male — resulting in classifiers which perform best for white males to the detriment of other subgroups. To minimise the risk of unfairness, AI systems must be designed in a bias-aware manner.

When biases are found, they can be corrected. Current approaches include imposing a fairness constraint on the loss function of the neural network \cite{dwork2012fairness}. The metric should be designed such that similar individuals are classified similarly — regardless of protected characteristics such as gender and race. Algorithmic bias is likely to be subject to increased scrutiny and legislation in the near future. Aside from the clear social and ethical issues raised, algorithmic bias poses a financial and legal risk which requires mitigation.

\begin{table}[!t] \renewcommand{\arraystretch}{1.3} \caption{\textsc{Demographic data of deepfake datasets. Datasets are ordered by their generation.}} \label{Demographic data of deepfake datasets}
\centering
\begin{tabular}{|m{4cm}|m{4cm}|}
\hline
 Dataset & Demographic Details  \\ \hline
 \multicolumn{2}{|c|}{First Generation} \\ \hline
DF-TIMIT \cite{korshunov2018deepfakesdftimit} & No demographic information provided. \\ \hline
UADFV \cite{korshunov2018deepfakesdftimit}& No demographic information provided. \\ \hline
\multicolumn{2}{|c|}{Second Generation} \\ \hline
Celeb-DF \cite{li2019celeb} & 59 celebrities in 590 real videos: 56.8\% male, 43.2\%  female. Age: under 30 (6.4\%), 30-39 (28.0\%), 40-49 (26.6\%), 50-59 (30.5\%), 60 and above (8.5\%). 88.1\% Caucasian, 6.8\% African American, and 5.1\% Asian.\\ \hline
DFDC Preview \cite{dolhansky2019deepfakedfdcpreview}&  74\% female, 26\% male. 68\% Caucasian, 20\% African-American, 9\% east-Asian, and 3\% south-Asian.\\ \hline
FaceForensics++ \cite{rossler2019faceforensics++xception} & Majority female, no further demographic information provided.\\ \hline

\multicolumn{2}{|c|}{Third Generation} \\ \hline
Deepfake Detection Challenge Dataset (DFDC) \cite{dolhansky2020deepfakedfdc} & The paper \cite{dolhansky2020deepfakedfdc} states that `the overall distribution of gender and appearance was balanced across all sets and video' but does not clarify how `appearance' is balanced. \\
\hline
\end{tabular}
\end{table}

Table I is a compilation of demographic and racial data from state of the art deepfake datasets. They are arranged by the generation to which we believe they belong. Notably, we consider FaceForensics++ to be a second generation dataset due to its size and inclusion of multiple deepfake generation methods. Due to the non-standardised reporting measures within the literature, all available details are noted. All of these datasets are lacking in vital demographic information.

These comparisons have clear limitations. The number of training examples that include these participants is not considered; subgroups may be underrepresented in the duration of videos taken and frames included within the dataset. In addition, the information on their participants varies between datasets. These factors must be considered when designing and presenting a new deepfake dataset and in particular when deploying a detection system in the real-world.

\section{Problem Analysis and Methodology}

\subsection{A New Framework for Classifying Deepfake Detection Approaches}

Li et al. \cite{li2019celeb} categorise existing deepfake detection methods into three groups: physiological, signal-level, and data-driven. These map onto the three sections outlined in the literature review. However, these categories do not consider the temporal aspects of a deepfake detection system. We propose an original framework for classifying deepfake detection techniques. This framework considers two principal components of a detection system: the method of feature extraction and the temporal aspect between frames. This approach allows for more effective comparison between the performance of detection systems. Table II shows the four categories of the framework.

\begin{table}[!t] \renewcommand{\arraystretch}{1.3} \caption{\textsc{Deepfake Framework: categorising deepfake detection approaches. Four distinct models, A, B, C, and D will be implemented.}} \label{tab:Deepfake Framework}

    \begin{tabular}{c|c|c|c|}
      \multicolumn{2}{c}{} & \multicolumn{2}{c}{Temporal Relationship}\\\cline{3-4}
      \multicolumn{1}{c}{} & &  Independent  & Dependent \\\cline{2-4}
      \multirow{2} * {Feature Extraction}  & Automatic &  A& B \\\cline{2-4}
      & Manual& C & D \\\cline{2-4}
    \end{tabular}
    \end{table}

Feature extraction refers to the manner in which the deepfake detection approach extracts features from the face of an individual. Manual feature extraction is where the detection system selects a specific feature, such as the difference in head pose estimates, that is supplied as a prior by the system's designer. Automatic feature extraction refers to a system where the type of feature selected is dependent on the training data used, such as the discriminating features learnt by a CNN. 

A detection system using an independent temporal relationship makes a classification on a frame-per-frame basis. If individual deepfake probabilities for frames are combined after calculation, the system still exhibits an independent temporal relationship. A temporally-dependent relationship takes into account the differences between frames through time. 

\subsection{Dataset Selection}

A subset of the FaceForensics++ \cite{rossler2019faceforensics++xception} dataset is used for the training, validation, and testing of each model presented. The dataset is chosen for its inclusion of numerous deepfake generation methods, the unconstrained nature of the videos, and the large size of the dataset. Furthermore, all videos are provided at different levels of compression — this more accurately represents videos shared on social media and allows for faster training of models. Combined, these characteristics allow for a better measure of the \emph{in-the-wild} performance of a deepfake detection system.

The performance of the implemented deepfake detection models are tested against three distinct deepfake generation methods found within the FaceForensics++ dataset: Deepfakes \cite{Deepfakes_2020}, FaceSwap \cite{faceswap_2020}, and Face2Face \cite{teadies2016face2face}. 1000 videos have been produced for each method alongside the original real videos. On social media, lossless video is rare. Hence, compressed versions of these videos are used throughout training, validation, and evaluation. These videos have been compressed by Rosser et al. \cite{rossler2019faceforensics++xception} using the H.264 codec with the constant rate quantization parameter set to 23 \cite{rossler2019faceforensics++xception}. This corresponds to the `High Quality' videos referenced in their work. 

From each video, 40 continuous frames of an individual's cropped face are extracted. Two videos in the deepfake video split did not meet this criterion which gives 3998 groups of 40 frames within the dataset. This results in 159,920 unique frames. Face extraction is performed using the face detector implemented within the \texttt{Dlib} library \cite{king2009dlib}. The extracted region from each frame is 1.3 times the size of the identified face as performed in \cite{rossler2019faceforensics++xception}. This results in a heavily pre-processed subset of the FaceForensics++ dataset, which will be referred to as TemporalFF++ from now onwards.

\begin{figure*}[h!]
\centering
\includegraphics[width = 0.7\textwidth]{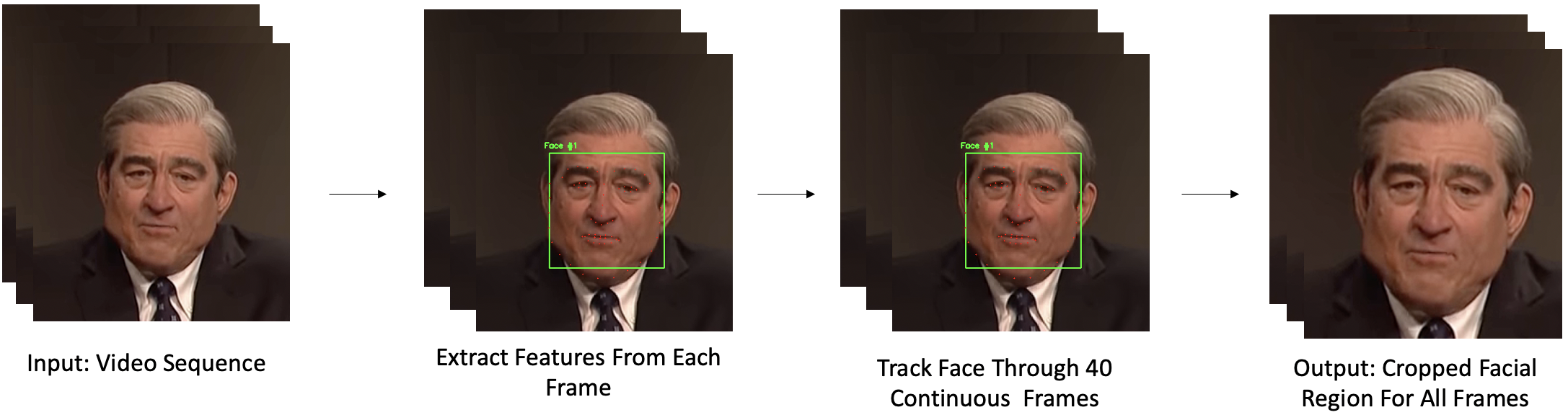}
\caption{Deepfake detection pipeline: pre-processing. The output is a 1.3x crop of the identified facial region.}
\end{figure*}

\subsection{The CNN-LSTM Architecture}

The use of a CNN-LSTM architecture allows for the creation of temporally-dependent models. The architecture is split into two separate neural networks: a convolutional neural network (CNN) and a long short term memory (LSTM). The CNN is used to extract features from input images. This process is repeated on $n$ frames ($n=40$ in this case) to form an $m \times n$ matrix where m is the feature vector. This matrix is passed to the LSTM for sequence analysis. 

LSTMs are a form of recurrent neural network (RNN). They have an internal state which enables information to persist within the network for an indefinite number of time steps \cite{bengio1994learningrnnproblem}. The internal state is used as short term memory while the weights are employed as a long term memory. However, Bengio et al. \cite{bengio1994learningrnnproblem} found that RNNs struggle to learn long-term dependencies when trained using gradient descent. Hochreiter \cite{hochreiter1999lstm} proposed a new architecture, the LSTM, which aimed to solve this problem. 

The LSTM is able to handle noisy inputs and learn long-term dependencies in excess of 1000 time steps \cite{hochreiter1999lstm}. It has been used to achieve state of the art performance in tasks such as neural machine translation \cite{wu2016googlelstm}. In investigating the effectiveness of temporal dependency for deepfake video detection, the LSTM allows contextual information from frames to be used in the classification of a video. 

\subsection{Deepfake Detection Pipeline}

The temporally-independent models  that we implement use a CNN architecture, whereas the temporally-dependent models use a combined CNN-LSTM architecture. Hence, the input for the CNN and the LSTM differ. The CNN requires a single frame as input whereas the LSTM requires a feature vector extracted from consecutive frames. To enable a fairer comparison between models, the same feature extractors will be used for the temporally-dependent and independent variant of each model.

The following outline summarises the steps performed upon the chosen subset of the FaceForensics++ before a prediction can be made. 

\begin{enumerate}
    \item Frame Extraction: The frame extraction script \cite{rossler2019faceforensics++xception} supplied with the FaceForensics++ dataset is adapted to extract cropped facial regions from each video only when a face is tracked through 40 consecutive frames. Once this has been achieved, the pixel values of each frame are normalised to the [0, 1] range.
    \item Feature extraction: The two variants of CNN architecture take the cropped image as input to produce either a probability value or a feature vector. The former is the case for the temporally-independent models whereas the temporally-dependent models use an intermediary feature vector as input to an LSTM. The LSTM makes a classification based on multiple temporally-related feature vectors. 
\end{enumerate}

Fig. 1 shows how this pipeline is used to pre-process the FaceForensics++ dataset to produce TemporalFF++.

\subsection{The Effect of Dataset Imbalance}

To produce realistic deepfakes, current synthesis processes require that only individuals with similar appearances are swapped. Dolhansky et al. \cite{dolhansky2019deepfakedfdcpreview} performed face swaps on those with similar appearances by considering their skin tone, facial hair, and use of glasses. Hence, a subgroup that is under-represented among real images is limited in the number of deepfakes that can be made using members of this subgroup compared to other subgroups. This feature of the deepfake generation process could further marginalise under-represented communities.

The following formalisation is an illustration of the effect dataset biases have on the number of deepfake videos that can be produced. Consider $n$ individuals all belonging to the same subgroup, each of which has $k$ real videos suitable to act as both target and source. The aim is to calculate the number of deepfake videos that can be produced from this subgroup of the dataset. Subsequently, the effect of changing the values of $n$ and $k$ will be explored.

This is a constrained permutation problem. Clearly, videos of individuals should not be swapped with other videos of themselves. This acts as a constraint on the problem. The number of videos that can be produced is equal to the number of ordered pairs that can be formed from $n$ individuals, each of which has $k$ videos. This is subject to the constraint that pairings can only be made between videos belonging to different individuals. The resulting Deepfake Production Equation is given below. 

\begin{equation}
 y = k^2n(n – 1)
 \label{DeepfakeProductionEquation}
\end{equation}

\begin{figure}[h!]
\centering
\includegraphics[width = 0.4\textwidth]{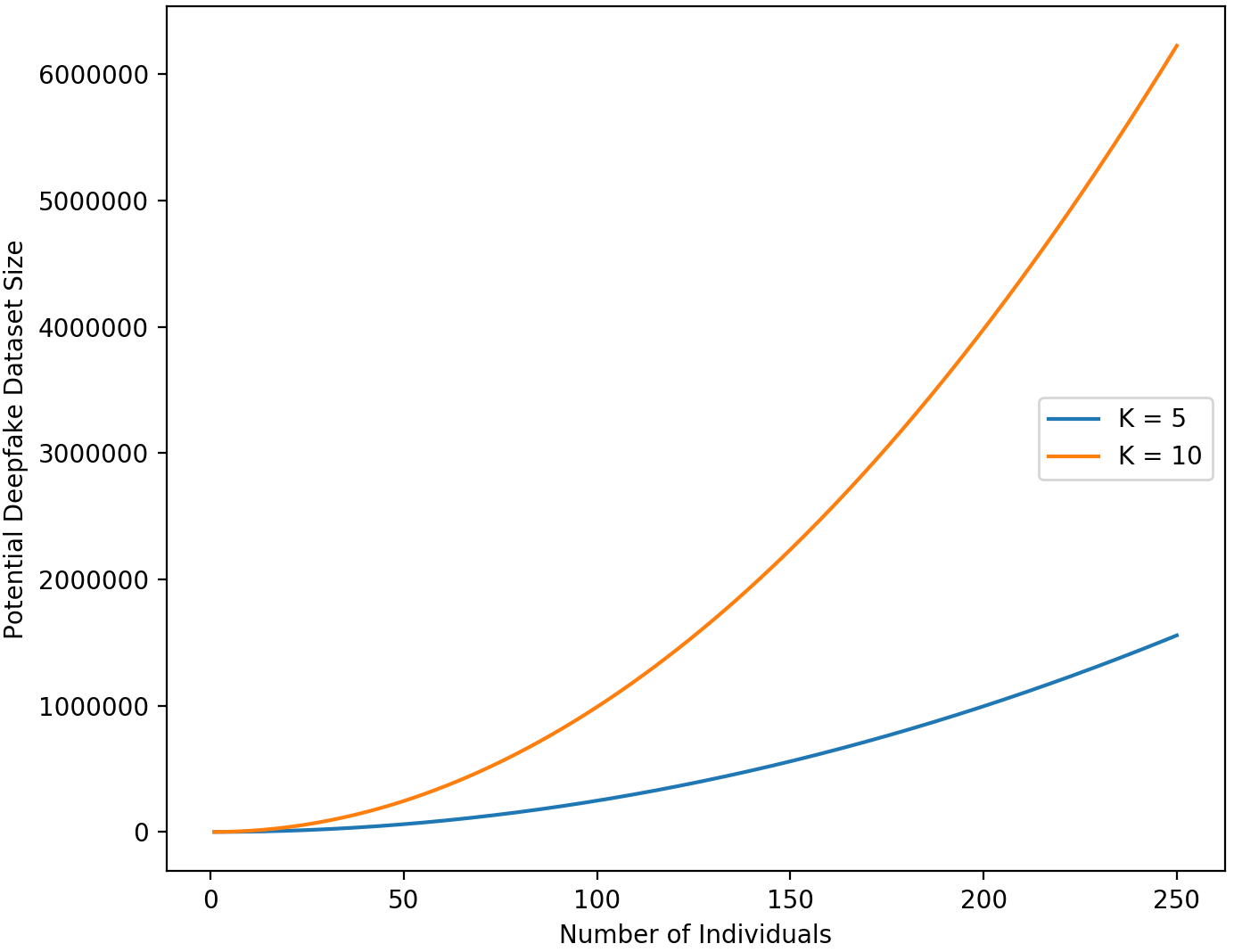}
\caption{Potential deepfake dataset size For n individuals.}
\end{figure}

Fig. 2 shows the effect of varying $n$ for different values of $k$. Clearly, as $k$ doubles, the number of potential deepfakes that can be generated quadruples. There is a quadratic relationship between these two variables. Hence, subgroups of individuals under-represented in real videos have far fewer other videos they can be swapped with — they have a low $k$ value. Hence, their under-representation in real videos will be amplified when considering the number of deepfake videos they may appear in. This finding further emphasises the need for explicit consideration of the diversity of individuals present within a dataset. This imbalance could be the cause of biases within a detection model trained on the dataset. In turn, this model could unfairly discriminate against individuals. 

\subsection{Training and Evaluation}

The three disjoint sets for training, validation, and evaluation proposed by Rossler et al.\cite{rossler2019faceforensics++xception} are followed. The categorical cross entropy loss is minimised during training. Training is stopped when either the validation accuracy of the LSTM reaches a 10-epoch plateau or 100 epochs are reached. The Adam optimiser \cite{kingma2014adam} is used with its default values \(\beta_{1}\) = 0.9, \(\beta_{2}\) = 0.999, \(\epsilon\) = \(e^{-8}\)).

Each deepfake detection model is trained and evaluated on all deepfake generation methods and real videos present in TemporalFF++. This results in a single probability of a video being a deepfake. To compensate for the resulting unbalanced dataset, incorrectly classified real examples are penalised more harshly using Keras's class weight functionality \cite{chollet2015keras}. 

\section{Design and Implementation}

\subsection{Implementation Environment}

All models were implemented and run using Python 3.8, Keras 2.2.4 \cite{chollet2015keras}, and Tensorflow 1.13.1 \cite{abadi2016tensorflow}. Open-CV \cite{opencv_library}, NumPy \cite{oliphantnumpy}, and MatPlotLib \cite{hunter2007matplotlib} were used for image processing, data manipulation, and data visualisation respectively. 

\subsection{Implemented Models}

Each implemented model has been adapted for the task of deepfake video detection, using transfer learning as a starting point for training. Auto and Warp each accept a single frame and output a probability of it being a deepfake. TAuto and TWarp are modified versions of these respective networks. They use a sequential CNN-LSTM  architecture to predict the probability of 40 continuous frames belonging to a deepfake video. The fit of these models into the Deepfake Framework is shown in Table III. 
\begin{table}[!t] \caption{\textsc{Classification of the four implemented models.}} \label{tab:Deepfake Detection Approaches Classification}
 \renewcommand{\arraystretch}{1.3}
\centering
    \setlength{\extrarowheight}{2pt}
    \begin{tabular}{c|c|m{2cm}|m{2cm}|}
      \multicolumn{2}{c}{} & \multicolumn{2}{c}{Temporal Relationship}\\\cline{3-4}
      \multicolumn{1}{c}{} &&  Independent  & Dependent \\\cline{2-4}
      \multirow{2} * {Feature Extraction}  & Automatic &  Auto & TAuto\\\cline{2-4}
      & Manual& Warp & TWarp \\\cline{2-4}
    \end{tabular}
    \end{table}

\begin{figure}[h!]
\centering
\includegraphics[height = 0.70\textheight]{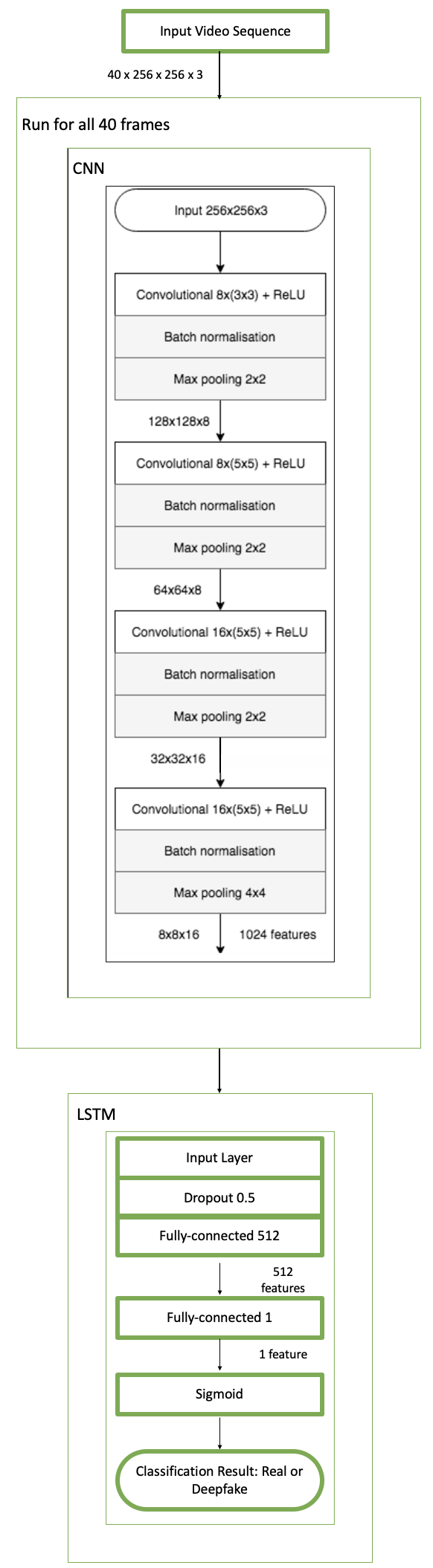}
\caption{The TAuto architecture. The diagram of the CNN is originally presented by Afchar et al. \cite{afchar2018mesonet}. Components added within this paper are marked in green. Transfer learning is used for the CNN while the LSTM is trained on TemporalFF++.}
\vspace{-5mm}
\end{figure}

We will now consider each implemented model in turn, noting all pertinent details. Fig. 3 shows a detailed view of TAuto and the CNN-LSTM architecture used. It is representative of the approach taken in designing all four models.

\subsubsection{Auto}

Auto employs the Meso-4 architecture of MesoNet proposed by Afchar et al.\cite{afchar2018mesonet}. It consists of a CNN architecture with four successive convolutional and max pooling layers. These are followed by a single dense layer which outputs the probability of a given image being a deepfake. The initial weights of the model are those generated from training on the private deepfake dataset created by Afchar et al. \cite{afchar2018mesonet}. From this starting point, it is trained on TemporalFF++.

\subsubsection{TAuto}

TAuto employs a CNN-LSTM architecture with the trained Auto network used as a feature extractor. This allow for fairer comparison between Auto and TAuto, isolating the inclusion of spatio-temporal information.  The final layers of Auto up to the pooling layer are removed, resulting in a 1024 element long feature vector being produced as output for each frame. An LSTM which accepts 40 of these feature vectors is constructed as the next part within the TAuto architecture. The LSTM consists of a 1024 units input layer, a dropout layer (rate=0.5), a 512 node dense layer, and finally a single sigmoid node to produce a classification. 

\subsubsection{Warp}

The Warp architecture is an adaptation of the pre-trained model designed by Li et al. \cite{li2018exposingfwa} to detect face warping artefacts produced by the deepfake generation process. It uses the CNN architecture of ResNet50 \cite{he2016deepresnet} to achieve this. It is trained on a private dataset of pristine images and images with face warping effects applied \cite{li2018exposingfwa}; the warped images are treated as positive examples of deepfakes despite the deepfake generation process not being explicitly applied. Hence, further training on the selected dataset is not performed; doing so would result in the model no longer taking advantage of its manually selected feature. This feature is not dataset specific but rather a model of the deepfake generation process, enabling a fair comparison between Warp and TWarp.

The weights from this pre-trained model are loaded and the layers up to the last pooling layer are removed. This results in a feature extractor which produces a 2048 element long feature vector. A 1024 node dense layer followed by a sigmoid output node are added after this layer of the network. The weights of the layers up to the feature extractor are frozen and these two layers are trained on TemporalFF++. This allows for the pre-selected feature to be preserved while ensuring that both Warp and TWarp are trained and evaluated on the same dataset.

\subsubsection{TWarp}

TWarp uses Warp for feature extraction and an LSTM for sequence analysis. The output from the final max pooling layer of Warp is taken as output; this produces a 2048 element feature vector for a given frame. The LSTM consists of a 2048 unit input layer, a dropout layer (rate=0.5), a 512 node dense layer, and a single sigmoid output node.

Input images are resized to 256x256 for Auto and 224x224 for Warp, reflecting the expected input image size of each network. Hence, the same image size is used for those models under comparison.

\section{Results}

\subsection{Results}

Table IV presents the area under the curve (AUC) of the precision-recall plot for each model. This measurement summarises the precision of a model for all possible values of recall. A high value represents both high precision and recall. In the context of deepfake video detection, precision is the percentage of deepfakes correctly identified out of all those identified as a deepfake. The recall is the percentage of all deepfakes present in the dataset that are correctly identified.

\begin{table}[!t]  \caption{\textsc{Recall-Precision AUC For Each Model.}} \label{tab:Recall-Precision AUC For Each Model}
\renewcommand{\arraystretch}{1.3}
\centering
\begin{tabular}{ |c|c|} 
\hline
Model   & AUC \\
\hline
Auto  & 0.940   \\
TAuto & 0.990  \\ 
Warp  & 0.856 \\
TWarp &  0.782   \\
\hline
\end{tabular}

\end{table}

Table V considers the precision of each model for a specific recall value. Clearly, TAuto exhibits higher precision than all other models for each of the three recall thresholds and a higher AUC. TWarp exhibits lower precision than Warp for all selected thresholds. The statistical significance of these differences must be considered before a causal relationship can be identified.

\begin{table}[!t]  \caption{\textsc{Precision for specified values of recall.}} \label{tab:Precision for specified values of recall}
\renewcommand{\arraystretch}{1.3}
\centering
\begin{tabular}{ |c|c|c|c|} 
\hline
Model/Recall   & 0.1 & 0.5 & 0.9\\
\hline
Auto  & 0.933	& 0.959 & 0.888\\ 
TAuto & 1 & 1 & 0.969  \\ 
Warp  & 0.933 & 0.933 & 0.704 \\
TWarp &  0.824 & 0.833 & 0.565  \\
\hline
\end{tabular}

\end{table}

\subsection{Statistical Significance \& Hypothesis Testing}

The temporally dependent TAuto produces the highest AUC value and highest precision for all three recall thresholds. However, in order to determine whether a specific model outperforms another model on a learning task, statistical significance is required. Statistical significance (p $<$ 0.05) allows us to determine whether a change in network architecture results in a significant change in performance. The same training, validation, and test data has been used for all trained models to ensure a fair comparison can be made. 

When comparing the results of two binary classifiers, their output can be viewed as paired nominal data. McNemar's test \cite{mcnemar1947note} has been recommended for comparing algorithms when evaluated once on a single test set \cite{dietterich1998mcnemar}. This is the case with deep, computationally-expensive models such as those presented in this paper. This test has been shown to have low type 1 error, meaning that the probability of it incorrectly detecting a statistically significant change is low \cite{dietterich1998mcnemar}. Hence, McNemar's test will be used to compare the results of the implemented models for statitistical significance.

McNemar's test is calculated from 2x2 contingency tables \cite{mcnemar1947note}. These tables contain counts of the paired distribution of correct and incorrect classifications for two classifiers under comparison. Calculation of McNemar's test statistic is performed using the \texttt{statsmodels} Python library \cite{seabold2010statsmodels}. Due to the limited number of conflicting outcomes within each contingency table, an exact binomial test is used to calculate the p-value in every case.

The two following hypotheses will be tested for using McNemar's test. These are tested for between each corresponding temporally dependent and independent model for every subset of the dataset.

\subsubsection*{H0} The two classifiers have the same distribution of errors on the test set.
\subsubsection*{H1} The two classifiers have a different distribution of errors on the test set.

 If p $<$ 0.05, H0 is rejected and the alternative hypothesis, H1, is accepted. Otherwise, H0, the null hypothesis, is accepted.

We do not know the prevalence of real videos and deepfake videos generated by each of the three methods present within TemporalFF++. Hence, statistical significance is calculated on the model's performance against each subset of the dataset. This results in eight null hypotheses and eight complementary alternative hypothesis: one for each comparison between two models on a single subset.

\subsection{Statistical Significance: Auto Vs. TAuto}

Auto and TAuto are models that are consistent in their method of automatic feature extraction but differ in how they interpret temporal information. This allows for the impact of the temporal relationship to be studied. Table VI presents the results of applying McNemar's test.

\begin{table}[!b] \renewcommand{\arraystretch}{1.3} \caption{\textsc{Statistical Significance: Auto and TAuto.}} \label{tab:Statistical Significance: Auto and TAuto}
\centering
\begin{tabular}{ |c|c| } 
\hline
  Evaluation Subset & p-value \\
\hline
Original  & 0.015    \\ 
Face2Face & 0.219  \\ 
FaceSwap  & 1    \\ 
Deepfakes & 1  \\ 
\hline

\end{tabular}
\end{table}

TAuto produces a statistically significant (p $<$ 0.05) increase in performance over Auto on the evaluation subset of real images. Hence, the alternative hypothesis is accepted: the two classifiers have a different distribution of errors on real images.

This shows that temporal information can improve classification performance of a deepfake classifier. Due to the imbalanced nature of deepfake detection, wherein the number of real videos is many orders of magnitude larger than the number of deepfake videos, an improvement in performance on real images offers a substantial improvement in expected real-world performance. 

The difference in performance for all other subsets is not statistically significant. Hence, the null hypothesis is accepted in each of these cases: the two classifiers have the same distribution of errors on each of these subsets.

\subsection{Statistical Significance: Warp Vs. TWarp}

Warp and TWarp both use the same manual method of feature extraction. The manually extracted feature is the face warping effect caused by the deepfake generation process. This comparison isolates the temporal component; the feature vector is followed by either a dense layer and output node or an LSTM and output node respectively. The statistical significance of the differences in their performance for each evaluation subset are detailed in Table VII below.

\begin{table}[!t] \renewcommand{\arraystretch}{1.3} \caption{\textsc{Statistical Significance: Warp and TWarp.}} \label{tab:Statistical Significance: Warp and TWarp}
\centering
\begin{tabular}{ |c|c| } 
\hline
  Evaluation Subset & p-value \\
\hline
Original  & 0.215    \\ 
Face2Face & 0.774  \\ 
FaceSwap  & 0.057    \\ 
Deepfakes & 1  \\ 
\hline

\end{tabular}
\end{table}

There are no statistically significant (p $<$ 0.05) differences in performance between Warp and TWarp for any of the four evaluation subsets. This suggests that temporal information does not result in a significant change in performance when using face warping features for a deepfake classifier. Intuitively, the variation of the face warping induced by the deepfake generation process over time is no more useful for detecting deepfakes than the same information that can be extracted from a single frame. Hence, the null hypothesis is accepted in all four cases — the classifiers do not have a different distribution of errors on the test set.

\subsection{Implications for Deepfake Video Detection}

A statistically significant (p $<$ 0.05) increase in performance  of TAuto over Auto was shown for the evaluation subset of real videos. This demonstrates that temporal information can improve the performance of deepfake detection models. Currently, a much larger number of real videos are published online rather than deepfake videos. Hence, models deployed in the real world could benefit from a substantial increase in performance by including temporal information as well as spatial information when classifying a video.

The precision of each model for three recall thresholds has been considered. In practice, the probability score output by a model could be interpreted differently based on the context in which it is deployed. For example, at a certain probability threshold, the video may be passed to an expert human reviewer to make a final judgement.

\subsection{Limitations}

FaceForensics++ is not balanced by gender and race. A deployed model must be evaluated against such a dataset to allow any biases against subgroups to be documented. A single model for each category of the Deepfake Framework has been implemented. However, there exists a large diversity of models within each class of the framework which require additional consideration. For example, the temporally dependent models only consider the use of LSTMs. Other neural network architectures such as the classical RNN could exhbit different performance characteristics.

\section{Conclusions}

In this paper, we presented a novel framework for classifying existing deepfake detection approaches. This framework was used to determine the effect of temporal dependence on deepfake video detection performance. Experimental results show that temporally-dependent approaches can outperform their temporally independent equivalents on real images to a statistically significant (p $<$ 0.05) degree. 

Further work must consider the social context in which deepfake detection systems are deployed — the threat posed by biased detection systems is clear. Studies to determine the prevalence of deepfakes in circulation on social media platforms would enable the real-world performance of deepfake detection models to be better quantified. We believe that the outlined framework and approach demonstrate an appropriate method to investigate the effect of temporal dependency and the method of feature extraction. 
\section*{acknowledgements}
We thank the University of York for supporting this research.

\bibliographystyle{IEEEtran}
\bibliography{IEEEfull,biblioR}

\begin{thebibliography}{10}
\providecommand{\url}[1]{#1}
\csname url@samestyle\endcsname
\providecommand{\newblock}{\relax}
\providecommand{\bibinfo}[2]{#2}
\providecommand{\BIBentrySTDinterwordspacing}{\spaceskip=0pt\relax}
\providecommand{\BIBentryALTinterwordstretchfactor}{4}
\providecommand{\BIBentryALTinterwordspacing}{\spaceskip=\fontdimen2\font plus
\BIBentryALTinterwordstretchfactor\fontdimen3\font minus
  \fontdimen4\font\relax}
\providecommand{\BIBforeignlanguage}[2]{{%
\expandafter\ifx\csname l@#1\endcsname\relax
\typeout{** WARNING: IEEEtran.bst: No hyphenation pattern has been}%
\typeout{** loaded for the language `#1'. Using the pattern for}%
\typeout{** the default language instead.}%
\else
\language=\csname l@#1\endcsname
\fi
#2}}
\providecommand{\BIBdecl}{\relax}
\BIBdecl

\bibitem{paris2019deepfakescheapfakes}
B.~Paris and J.~Donovan, ``Deepfakes and cheap fakes,'' \emph{United States of
  America: Data \& Society}, 2019.

\bibitem{doublicat}
\BIBentryALTinterwordspacing
``doublicat| face swap in gifs\_2020,'' 2020. [Online]. Available:
  \url{https://doublicat.com}
\BIBentrySTDinterwordspacing

\bibitem{schroff2015facenet}
F.~Schroff, D.~Kalenichenko, and J.~Philbin, ``Facenet: A unified embedding for
  face recognition and clustering,'' in \emph{Proceedings of the IEEE
  conference on computer vision and pattern recognition}, 2015, pp. 815--823.

\bibitem{korshunov2018deepfakesdftimit}
P.~Korshunov and S.~Marcel, ``Deepfakes: a new threat to face recognition?
  assessment and detection,'' \emph{arXiv preprint arXiv:1812.08685}, 2018.

\bibitem{vaccari2020deepfakes}
C.~Vaccari and A.~Chadwick, ``Deepfakes and disinformation: exploring the
  impact of synthetic political video on deception, uncertainty, and trust in
  news,'' \emph{Social Media+ Society}, vol.~6, no.~1, p. 2056305120903408,
  2020.

\bibitem{chesneyposneg2019deep}
B.~Chesney and D.~Citron, ``Deep fakes: a looming challenge for privacy,
  democracy, and national security,'' \emph{Calif. L. Rev.}, vol. 107, p. 1753,
  2019.

\bibitem{dolhansky2020deepfakedfdc}
B.~Dolhansky, J.~Bitton, B.~Pflaum, J.~Lu, R.~Howes, M.~Wang, and C.~C. Ferrer,
  ``The deepfake detection challenge dataset,'' 2020.

\bibitem{li2019celeb}
Y.~Li, X.~Yang, P.~Sun, H.~Qi, and S.~Lyu, ``Celeb-df: A new dataset for
  deepfake forensics,'' \emph{arXiv preprint arXiv:1909.12962}, 2019.

\bibitem{nirkin2019fsgan}
Y.~Nirkin, Y.~Keller, and T.~Hassner, ``Fsgan: Subject agnostic face swapping
  and reenactment,'' in \emph{Proceedings of the IEEE International Conference
  on Computer Vision}, 2019, pp. 7184--7193.

\bibitem{suwajanakorn2017synthesizingobama}
S.~Suwajanakorn, S.~M. Seitz, and I.~Kemelmacher-Shlizerman, ``Synthesizing
  obama: learning lip sync from audio,'' \emph{ACM Transactions on Graphics
  (TOG)}, vol.~36, no.~4, pp. 1--13, 2017.

\bibitem{teadies2016face2face}
J.~Thies, M.~Zollhofer, M.~Stamminger, C.~Theobalt, and M.~Nie{\ss}ner,
  ``Face2face: Real-time face capture and reenactment of rgb videos,'' in
  \emph{Proceedings of the IEEE conference on computer vision and pattern
  recognition}, 2016, pp. 2387--2395.

\bibitem{rossler2019faceforensics++xception}
A.~Rossler, D.~Cozzolino, L.~Verdoliva, C.~Riess, J.~Thies, and M.~Nie{\ss}ner,
  ``Faceforensics++: Learning to detect manipulated facial images,'' in
  \emph{Proceedings of the IEEE International Conference on Computer Vision},
  2019, pp. 1--11.

\bibitem{petrov2020deepfacelab}
I.~Petrov, D.~Gao, N.~Chervoniy, K.~Liu, S.~Marangonda, C.~Um{\'e}, J.~Jiang,
  L.~RP, S.~Zhang, P.~Wu \emph{et~al.}, ``Deepfacelab: A simple, flexible and
  extensible face swapping framework,'' \emph{arXiv preprint arXiv:2005.05535},
  2020.

\bibitem{yang2019exposingheadpose}
X.~Yang, Y.~Li, and S.~Lyu, ``Exposing deep fakes using inconsistent head
  poses,'' in \emph{ICASSP 2019-2019 IEEE International Conference on
  Acoustics, Speech and Signal Processing (ICASSP)}.\hskip 1em plus 0.5em minus
  0.4em\relax IEEE, 2019, pp. 8261--8265.

\bibitem{li2018exposingfwa}
Y.~Li and S.~Lyu, ``Exposing deepfake videos by detecting face warping
  artifacts,'' \emph{arXiv preprint arXiv:1811.00656}, 2018.

\bibitem{he2016deepresnet}
K.~He, X.~Zhang, S.~Ren, and J.~Sun, ``Deep residual learning for image
  recognition,'' in \emph{Proceedings of the IEEE conference on computer vision
  and pattern recognition}, 2016, pp. 770--778.

\bibitem{li2018ictueyes}
Y.~Li, M.-C. Chang, and S.~Lyu, ``In ictu oculi: Exposing ai created fake
  videos by detecting eye blinking,'' in \emph{2018 IEEE International Workshop
  on Information Forensics and Security (WIFS)}.\hskip 1em plus 0.5em minus
  0.4em\relax IEEE, 2018, pp. 1--7.

\bibitem{deng2009imagenet}
J.~Deng, W.~Dong, R.~Socher, L.-J. Li, K.~Li, and L.~Fei-Fei, ``Imagenet: A
  large-scale hierarchical image database,'' in \emph{2009 IEEE conference on
  computer vision and pattern recognition}.\hskip 1em plus 0.5em minus
  0.4em\relax Ieee, 2009, pp. 248--255.

\bibitem{chollet2017xception}
F.~Chollet, ``Xception: Deep learning with depthwise separable convolutions,''
  in \emph{Proceedings of the IEEE conference on computer vision and pattern
  recognition}, 2017, pp. 1251--1258.

\bibitem{tursman2020towardssocialvideoveri}
E.~Tursman, M.~George, S.~Kamara, and J.~Tompkin, ``Towards untrusted social
  video verification to combat deepfakes via face geometry consistency,'' in
  \emph{Proceedings of the IEEE/CVF Conference on Computer Vision and Pattern
  Recognition Workshops}, 2020, pp. 654--655.

\bibitem{buolamwini2018gender}
J.~Buolamwini and T.~Gebru, ``Gender shades: Intersectional accuracy
  disparities in commercial gender classification,'' in \emph{Conference on
  fairness, accountability and transparency}, 2018, pp. 77--91.

\bibitem{dwork2012fairness}
C.~Dwork, M.~Hardt, T.~Pitassi, O.~Reingold, and R.~Zemel, ``Fairness through
  awareness,'' in \emph{Proceedings of the 3rd innovations in theoretical
  computer science conference}, 2012, pp. 214--226.

\bibitem{dolhansky2019deepfakedfdcpreview}
B.~Dolhansky, R.~Howes, B.~Pflaum, N.~Baram, and C.~C. Ferrer, ``The deepfake
  detection challenge (dfdc) preview dataset,'' \emph{arXiv preprint
  arXiv:1910.08854}, 2019.

\bibitem{Deepfakes_2020}
\BIBentryALTinterwordspacing
``Deepfakes,'' 2020. [Online]. Available:
  \url{https://github.com/deepfakes/faceswap}
\BIBentrySTDinterwordspacing

\bibitem{faceswap_2020}
\BIBentryALTinterwordspacing
``Faceswap,'' 2020. [Online]. Available:
  \url{https://github.com/MarekKowalski/FaceSwap/}
\BIBentrySTDinterwordspacing

\bibitem{king2009dlib}
D.~E. King, ``Dlib-ml: A machine learning toolkit,'' \emph{The Journal of
  Machine Learning Research}, vol.~10, pp. 1755--1758, 2009.

\bibitem{bengio1994learningrnnproblem}
Y.~Bengio, P.~Simard, and P.~Frasconi, ``Learning long-term dependencies with
  gradient descent is difficult,'' \emph{IEEE transactions on neural networks},
  vol.~5, no.~2, pp. 157--166, 1994.

\bibitem{hochreiter1999lstm}
S.~Hochreiter and J.~Schmidhuber, ``Long short-term memory,'' \emph{Neural
  computation}, vol.~9, no.~8, pp. 1735--1780, 1997.

\bibitem{wu2016googlelstm}
Y.~Wu, M.~Schuster, Z.~Chen, Q.~V. Le, M.~Norouzi, W.~Macherey, M.~Krikun,
  Y.~Cao, Q.~Gao, K.~Macherey \emph{et~al.}, ``Google's neural machine
  translation system: Bridging the gap between human and machine translation,''
  \emph{arXiv preprint arXiv:1609.08144}, 2016.

\bibitem{kingma2014adam}
D.~P. Kingma and J.~Ba, ``Adam: A method for stochastic optimization,''
  \emph{arXiv preprint arXiv:1412.6980}, 2014.

\bibitem{chollet2015keras}
F.~Chollet \emph{et~al.}, ``Keras,'' \url{https://keras.io}, 2015.

\bibitem{abadi2016tensorflow}
M.~Abadi, P.~Barham, J.~Chen, Z.~Chen, A.~Davis, J.~Dean, M.~Devin,
  S.~Ghemawat, G.~Irving, M.~Isard \emph{et~al.}, ``Tensorflow: A system for
  large-scale machine learning,'' in \emph{12th $\{$USENIX$\}$ Symposium on
  Operating Systems Design and Implementation ($\{$OSDI$\}$ 16)}, 2016, pp.
  265--283.

\bibitem{opencv_library}
G.~Bradski, ``{The OpenCV Library},'' \emph{Dr. Dobb's Journal of Software
  Tools}, 2000.

\bibitem{oliphantnumpy}
T.~E. Oliphant, \emph{A guide to NumPy}.\hskip 1em plus 0.5em minus 0.4em\relax
  Trelgol Publishing USA, 2006, vol.~1.

\bibitem{hunter2007matplotlib}
J.~D. Hunter, ``Matplotlib: A 2d graphics environment,'' \emph{Computing in
  science \& engineering}, vol.~9, no.~3, pp. 90--95, 2007.

\bibitem{afchar2018mesonet}
D.~Afchar, V.~Nozick, J.~Yamagishi, and I.~Echizen, ``Mesonet: a compact facial
  video forgery detection network,'' in \emph{2018 IEEE International Workshop
  on Information Forensics and Security (WIFS)}.\hskip 1em plus 0.5em minus
  0.4em\relax IEEE, 2018, pp. 1--7.

\bibitem{mcnemar1947note}
Q.~McNemar, ``Note on the sampling error of the difference between correlated
  proportions or percentages,'' \emph{Psychometrika}, vol.~12, no.~2, pp.
  153--157, 1947.

\bibitem{dietterich1998mcnemar}
T.~G. Dietterich, ``Approximate statistical tests for comparing supervised
  classification learning algorithms,'' \emph{Neural computation}, vol.~10,
  no.~7, pp. 1895--1923, 1998.

\bibitem{seabold2010statsmodels}
S.~Seabold and J.~Perktold, ``Statsmodels: Econometric and statistical modeling
  with python,'' in \emph{Proceedings of the 9th Python in Science Conference},
  vol.~57.\hskip 1em plus 0.5em minus 0.4em\relax Austin, TX, 2010, p.~61.

\end{thebibliography}

\end{document}